\documentclass{article}

\usepackage[preprint]{neurips_2020}

\usepackage[utf8]{inputenc} % allow utf-8 input
\usepackage[T1]{fontenc}    % use 8-bit T1 fonts
\usepackage{url}            % simple URL typesetting
\usepackage{booktabs}       % professional-quality tables
\usepackage{amsfonts}       % blackboard math symbols
\usepackage{nicefrac}       % compact symbols for 1/2, etc.
\usepackage{microtype}      % microtypography
\usepackage{graphicx}
\usepackage{multirow}
\usepackage{xcolor}

\usepackage{amsmath,amsfonts,amssymb,dsfont}
\usepackage{hyperref}       % hyperlinks

\def\e{\varepsilon}
\def\lip{\mathrm{lip}}

\def\R{\mathbb{R}}

\def\D{\mathcal{D}}
\def\E{\mathcal{E}}

\title{Adversarial Examples Detection and Analysis with Layer-wise Autoencoders}
\author{%
	Bartosz Wójcik \\
	Jagiellonian University\\
	\And
	Paweł Morawiecki \\
	Institute of Computer Science\\
	Polish Academy of Sciences \\
	\And
	Marek Śmieja \\
	Jagiellonian University\\
	\And
	Tomasz Krzyżek \\
	Jagiellonian University\\
	\And
	Przemysław Spurek \\
	Jagiellonian University\\
	\And
	Jacek Tabor \\
	Jagiellonian University\\
}

\begin{document}
\maketitle

\begin{abstract}
We present a mechanism for detecting adversarial examples based on data representations taken from the hidden layers of the target network. For this purpose, we train individual autoencoders at intermediate layers of the target network. This allows us to describe the manifold of true data and, in consequence, decide whether a given example has the same characteristics as true data. It also gives us insight into the behavior of adversarial examples and their flow through the layers of a deep neural network. Experimental results show that our method outperforms the state of the art in supervised and unsupervised settings.
\end{abstract}

%%%%%%%%%%%%%%%%%%%%%%%%%%%%%%%%%%%%%%%%%%%%%%%%%%%%%%%%%%%%%%%%%%%%%%%%%%%%%%%

\section{Introduction}
\label{submission}

Deep neural networks have shown impressive performance on various machine learning tasks including object detection \citep{zhao2019object}, speech recognition \citep{amodei2016deep}, image classification \citep{he2016deep}, etc. While these models are usually robust to random noise, their performance can dramatically deteriorate under adversarial perturbations, i.e., small changes of the input which are imperceptible to humans, but mislead the model to output wrong predictions \citep{szegedy2013intriguing, goodfellow2014explaining}. This phenomenon can disqualify a model from applications such as autonomous cars or banking systems, where security is a priority \citep{sitawarin2018darts, grosse2017adversarial}.

%related works- defenses
Several methods have been proposed to defend the deep learning models against adversarial attacks. One approach relies on adding adversarial examples to the training stage, which makes the model robust to many (but not all) adversarial attacks \citep{madry2017towards}. 
To give formal guarantees that no adversarial perturbation within a given range fools a neural network, more computationally demanding provable defenses are used. These methods employ either integer programming approaches \citep{lomuscio2017approach, xiao2018training}, Satisfiability
Modulo Theories (SMT) solvers \citep{carlini2017provably, ehlers2017formal} or computing an approximation to the adversarial polytope \citep{zhang2018efficient, morawiecki2019fast}. All of the aforementioned approaches involve special training procedures and, in consequence, they cannot be used when the target neural network is fixed (we are not allowed to modify or retrain it from scratch). 

%related works - detection
The other line of research, which is considered in this paper, relies on introducing auxiliary mechanisms to detect whether the input can be seen as an adversarial example without modifying the target model. Most methods employ a supervised approach and train a classifier to discriminate normal samples from adversarial examples \citep{hendrycks2016baseline}. In \citep{lee2018simple}, the authors propose to estimate the class-conditional Gaussian distribution in each layer and use the Mahalanobis distance to discriminate adversarial examples. A different approach is to consider the unsupervised detection of abnormal samples.
In \citep{xu2017feature}, adversarial examples are detected by comparing the model’s predictions on a given input with its predictions on a squeezed version of the input. The authors of \citep{yang2019ml} propose a detection mechanism based on the observation that the feature attribution map of an adversarial example near the boundary always differs from that of the corresponding original example. In \citep{roth2019odds}, the authors introduce a statistical test based on the change of feature representations and log-odds under noise. The authors of \citep{samangouei2018defense} use GANs to model the distribution of unperturbed images and, in consequence, find a close output to a given image which does not contain the adversarial changes. An interesting analysis in \citep{deep_knn}, which partially inspired the presented paper, shows how the representation of an adversarial example changes through neural network layers.

The primary goal of this paper is to characterize and analyze the behavior of the adversarial examples, see Section \ref{sec:adversarial_dynamics} for the formulation of our hypotheses and  Section \ref{sec:analysis} for experimental evidence. Our first observation shows that as we increase the perturbation, the movement of adversarial examples from the true data manifold is stronger than their movement within that manifold. The second observation is that the deviation of adversarial examples from the normal data manifold can be observed in the hidden layers of the target network and the deepest layers are the most discriminative. Our analysis extends the recent works concerning adversarial attacks \citep{lee2018simple, deep_knn, pidhorskyi2018generative}.

\begin{figure}[!ht] 
\begin{center} 
 \centerline{\includegraphics[width=0.8\columnwidth]{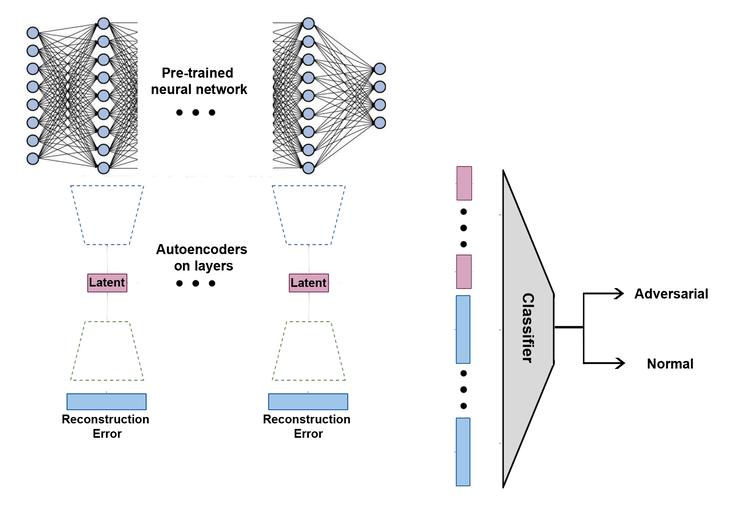}}

 \caption{The scheme of the proposed detection method. We train individual autoencoders on hidden representations of the pre-trained target network to describe the manifold of true data. Features from the autoencoders are used by an external classifier to detect the adversarial examples.}
\label{fig:detection_method} 
\end{center} 
\end{figure} 

Our second contribution is a method for detecting adversarial examples based on data representation taken from the hidden layers of the target network, see Figure \ref{fig:detection_method}. For this purpose, we train individual generative autoencoders at hidden layers of the target network. This allows us to describe the original data manifold and helps to decide whether a given example has the same characteristic as the actual data. We stress that our solution does not require retraining of the target network, so it can be applied to existing models without modifying them.

Experiments performed on the ResNet and DenseNet architectures show that the implementation of our method gives the highest detection rate among recent detection methods, see Table \ref{tab:results1}. Moreover, our analysis sheds a new light on the understanding of the behavior of adversarial examples.

%%%%%%%%%%%%%%%%%%%%%%%%%%%%%%%%%%%%%%%

\section{Detection and Analysis of Adversarial Examples}
\label{sec:adversarial_dynamics}

Before we describe the detection method, let us first discuss our observations regarding the behavior of the adversarial examples in deep neural networks. We start with the basics of adversarial examples and generative models. Next, we formalize our research hypotheses, which describe the behavior of adversarial examples in neural network layers. This analysis naturally leads us to the detection algorithm.

\subsection{Preliminaries}

{\bf Neural network.} We consider a neural network $F: \R^D \to \R^K$, which maps the input data $X \subset \R^D$ to the logit outputs for the $K$-class classification problem. The network is composed of $L$ layers $h_l$:
$$
z_l = h_l(z_{l-1}), \text{ for } l=1,\ldots,L.
$$ 
A final classification is obtained by applying the softmax function $S$ to the logits $z_L$.

{\bf Adversarial attack.} Typical classification networks tend to have strong predictions on unseen data. For example, if a network is trained on the classes "1-9" of the MNIST data set, it will also provide strong predictions for elements of class "0". In other words, a basic classification network cannot reliably estimate the uncertainty of the predictions, which, in particular, makes it vulnerable to adversarial attacks.

In adversarial attacks, the attacker adds a small perturbation $\delta$ to the input $x$, such that a neural network gives wrong prediction, i.e., 
$$
\arg \max_i S(F(x+\delta))_i \neq y_{true}, 
$$
where $y_{true}$ denotes the true class label of $x$. The perturbed input $x + \delta$, which causes the network to misclassify $x$, is called an \textit{adversarial example}. For most adversarial attacks we assume that the perturbation magnitude $\delta$ cannot be bigger than a fixed value $\e$ measured by the distance metric $d$.

\textbf{Autoencoders.} According to the manifold hypothesis, high-dimensional data tend to lie in a low-dimensional manifold \citep{fefferman2013testing}. One approach for describing a data manifold is to use the autoencoder model. The classical autoencoder (AE) consists of the encoder $\E$ and the decoder $\D$. The encoder transports the input data to a generally lower-dimensional latent space $\E: \R^d \to Z = \R^n$, whereas the decoder transforms the latent $Z$ back to the original space $\D:Z \to \R^d$. We search for $\E$ and $\D$ such that the reconstruction error 
$$
Error(X;\E,\D) = \sum_{x \in X} \|x - \D(\E(x))\|^2
$$ 
 is minimized, where $X \subset \R^d$ denotes the training data. Intuitively, the reconstruction error describes a distance of a given sample from the manifold.

If we want to additionally model a probability distribution of data on that manifold, we can define a prior distribution $f$ (typically Gaussian) in the latent space and optimize the distance of $\E(x)$ from the prior, which is the basic idea of the autoencoder-based generative models \citep{tabor2018cramer}. In the case of the Variational AutoEncoder (VAE) \citep{VAE}, the discrepancy between $\E(x)$ and $f$ is measured via the Kullback-Leibler divergence. In the Wasserstein autoencoder (WAE) \citep{WAE}, we use the Wasserstein distance (implemented either as the MMD or GAN loss). 

\subsection{Adversarial examples dynamics}

The main idea pursued in this paper is that adversarial examples have a different distribution from normal data \citep{grosse2017statistical}. To quantify this discrepancy, we formulate two hypotheses. The first one states that the increase of the perturbation magnitude pushes adversarial examples further from the manifold of normal data. The second one states that the distribution disagreement cannot be easily detected in the original data representation (inputs), but can be observed in the hidden layers of the target network. The hypotheses verification is the subject of Section \ref{sec:analysis}.

{\bf Influence of perturbation magnitude.} We assume that $M$ is the manifold of normal data $X \subset \R^D$. Let us consider the $\e$-bounded attack, where the attacker is allowed to perturb the input data point $x$ by a maximal value $\e$. To illustrate the dependence of an adversarial example on the maximal perturbation $\e$, we consider the curve:
$$
\gamma: \e \to x_\e,
$$
which for a given $\e$ and a fixed $x \in X$ returns an adversarial example $x_\e = x + \delta$, where $\delta \in [\e, \e]^D$. In other words, we look at the trajectory of the adversarial example with the increasing magnitude of the perturbation.

From this point of view, we can decompose the curve $\gamma$ into the component from data manifold $M$ and its orthogonal complement as:
$$
\gamma=\gamma_M+\gamma_{\perp M},
$$
where 
$$
\gamma_M(\e) \in M, \gamma_{\perp M}(\e)-\gamma_M(\e) \perp 
T_{\gamma_M(\e)}M.
$$
and $T_{\gamma_M(\e)}$ is a tangent space to $M$ at the point $\gamma(\e)$. Generally, we choose $\gamma_M(\e)$ to be the nearest point in $M$ from $\gamma(\e)$.

The main question is: {\em which of these two components has a stronger influence on the construction of adversarial examples?}

{\em Adversarial dynamics hypothesis:} The movement perpendicular to the manifold $M$ of true data is dominant for the construction of adversarial examples, i.e., the deviation of $\gamma_{\perp M}(\e)$ is higher than the deviation of $\gamma_{M}(\e)$, with the increasing value of the maximal perturbation $\e$.

In the context of generative AEs:
\begin{itemize}
    \item The movement along $\gamma_{\perp M}$ is quantified by the reconstruction error. This coincides with a discrepancy between a given sample and the data manifold.
    \item The movement within $\gamma_{M}$ is quantified by the norm in the latent space. It shows whether a data point is generated from the data distribution on that manifold.
\end{itemize}

{\bf Reconstruction error propagation through layers.} According to the previous hypothesis, adversarial examples move far apart from data manifold as the perturbation magnitude increases. However, since the difference between clean and adversarial examples is very small, it is usually difficult to detect adversarial attacks using their initial representation (inputs to the network). To overcome this difficulty, we study the hidden layers of the network. Since each layer gives a different representation of the same data, we can now ask: {\em which layer has the highest discriminative power in the context of adversarial examples?}

Observe that the network is, in fact, a composition of nonlinear maps
$$
F=h_L \circ \ldots \circ h_1.
$$
Since, in practice, each of these maps has the Lipschitz constant $\lip(h_l)$ greater than $1$, we can argue that a small movement in the input space may result in a very large representation distance in the last layers. Moreover, in high-dimensional spaces we usually have
$$
\lip(F) \approx \lip(h_1) \cdot \ldots \cdot \lip(h_L). 
$$
Consequently, with a very small change in the input, we expect high divergence from the manifold of normal data, which is the main point of our second hypothesis:

\emph{Layers dynamics hypothesis:} The divergence of adversarial examples from the manifold (and corresponding distribution) of normal data is more evident in the consecutive hidden layers than in the initial representation. 

Exploration of this hypothesis is also motivated by other works where information is gathered and combined from the hidden layers. In \citep{deep_knn}, an adversarial example is compared to the closest neighbor from the training set on a layer-by-layer basis; in \citep{lee2018simple}, the Mahalanobis distance is calculated between the adversarial example and the training set, also across the layers.

\begin{figure*}[!ht] 
 \centering
 \centerline{\includegraphics[height=9cm]{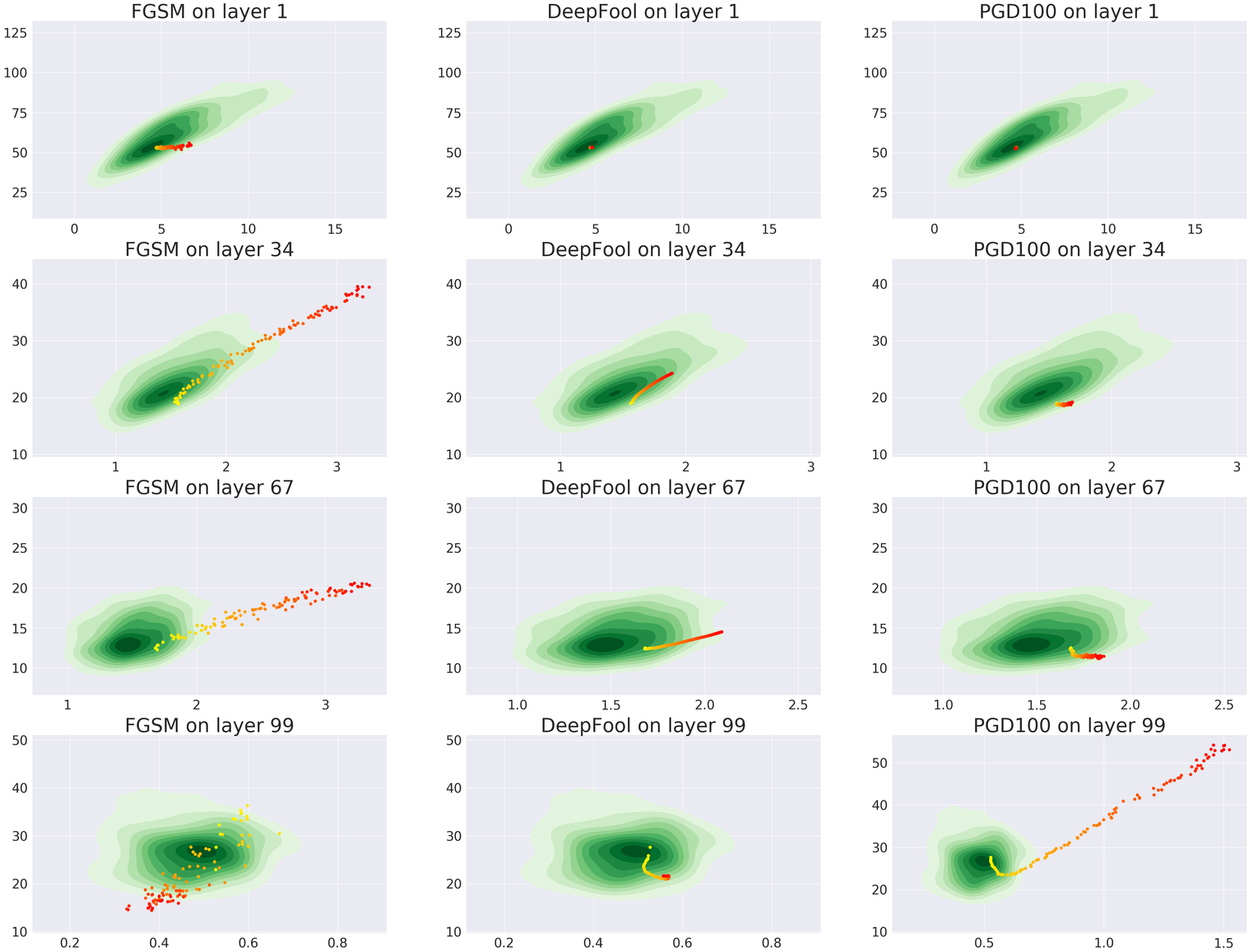}}
 \caption{Impact of adversarial perturbations on the position of an example in 2D space (reconstruction error in the x-axis and latent $L_2$ norm in the y-axis). The starting point is a single normal example without any perturbations (visualized as a light yellow dot). Then, we linearly increase the perturbation until it reaches the value of $2\epsilon$ (dark red). The $\epsilon$ values for a given network architecture/dataset are the same as in \citep{lee2018simple}. As a point of reference, we provide the kernel density estimation (green) of normal examples from the test set. Each column of sub-figures refers to a different attack and each row of sub-figures refers to a different layer. The experiments shown here are for the Densenet architecture and the CIFAR-100 dataset. The crucial observation is that adversarial examples significantly move away from the data manifold as the perturbation magnitude increases (movement in x-axis).}
\label{fig:adv_epsilon_impact} 
\end{figure*}

\subsection{Detection Mechanism}

Based on the aforementioned hypotheses, we formulate a mechanism for detecting abnormal samples, depicted in Figure \ref{fig:detection_method}. 

To describe the manifold of normal data $X$, we train individual generative autoencoders at consecutive layers of the network $F$. To reduce training and testing time, we use only selected layers of the network. We emphasize that only the normal data is used for the autoencoder training and the target network weights remain unmodified.

As explained, the reconstruction error and the latent norm can be seen as two factors that measure the discrepancy from data manifold and divergence from the corresponding distribution. In consequence, the classification system, which is responsible for detecting abnormal samples, is trained on these features. More precisely, we collect these two features from all AEs and pass them to a given classifier. In the experiments, we also investigate the situation when AEs are represented by the full latent vectors instead of the reconstruction error and the latent norm. This alternative representation can be more meaningful, but, one the other hand, the classification model is more time consuming to train and test on that high-dimensional space.

Our method can be applied to two real-life scenarios. In the first one, we assume that our system is targeted for a specific attack type. In this case, we train a (fully supervised) classifier on normal and adversarial examples. In a more realistic scenario, where the type of an adversarial attack is unknown, we cannot train the final classifier in the supervised way as we lack the second (adversarial) class examples. In this situation, we build a one-class classifier, which is trained only on normal data. In the test stage, we verify whether a sample belongs to data or not.

\section{Analysis of Adversarial Examples Dynamics} \label{sec:analysis}

In this section, we experimentally verify the hypotheses formulated in Section \ref{sec:adversarial_dynamics}. For this purpose, we use two modern CNN architectures: ResNet and DenseNet for three classification tasks: CIFAR-10, CIFAR-100, and SVHN. We consider five different attacks: FGSM, BIM, DeepFool, CW ($L_{2}$) and PGD ($L_{\inf}$, 100 iterations). To implement our detection mechanism, we train WAE-MMD autoencoders on the representations taken from selected hidden layers using training data. More details on training parameters and the experiment setup are given in Section \ref{sec:experiments}, where we present the results for the detection problem.

{\bf Deviation from data manifold (Hypothesis 1).} In the first experiment, we investigate the influence of the perturbation magnitude on the location of adversarial examples in the normal data distribution. For this purpose, we generate adversarial examples with increasing values of $\e$ for three types of attacks\footnote{As the CW attack is optimization-based, it is not straightforward to manipulate and fix $\e$ for the needs of this experiment}. We visualize the experiment in 2D space, where the x-axis is the reconstruction error of the WAE autoencoder and the y-axis is the $L_2$ norm in the latent space. The experiment is performed on the CIFAR-10 dataset and the ResNet architecture. We provide figures for all other dataset-architecture combinations in Supplementary Material, Section B.

It is evident from Figure \ref{fig:adv_epsilon_impact} that the curve $\gamma(\e)$ gradually moves away from the distribution of normal data. The movement in the direction perpendicular to the data manifold (x-axis) is stronger than within the manifold (y-axis). This discrepancy is most evident in layers 34 and 99 (second and fourth rows). It partially confirms that the movement in the direction perpendicular to data manifold is essential.

% To support our analysis in a more quantitative way, we use the random forest classifier as the final classifier in our detection system. It is trained on the composition of the reconstruction error and the latent norm from each WAE. As a result, we obtain the importance of each feature. We aggregate these indicators over the layers to get the summarized importance of the reconstruction error and the latent norm. The reconstruction error is significantly more important than the latent norm for the detection of abnormal samples (see Supplementary Material for figures illustrating feature importances for different dataset-architecture settings). This confirms our hypothesis that adversarial examples deviate from the data manifold more than from the distribution on that manifold. %In Supplementary Material we provide figures illustrating feature importances for different dataset-architecture settings.

To support our analysis in a more quantitative way, we assess the feature importance in the detection context. For this purpose, we consider supervised and unsupervised settings. In the first case, we train a random forest classifier on the composition of the reconstruction error and the latent norm from each WAE. As a result, we obtain the importance of each feature, which are aggregated over the layers to get the summarized importance of the reconstruction error and the latent norm. In the unsupervised case, we report the performance of one-class classifier (implemented as the Isolation Forest) trained on either the reconstruction error only, the latent norm only, or on both features. The results presented in Supplementary Material, Section D show that the reconstruction error is significantly more important than the latent norm for the detection of abnormal samples. We also conduct a similar ablation study for the supervised classifier, where the detector is trained in a few scenarios, each with a different set of features (see Supplementary Material, Section E). These findings confirm our hypothesis that adversarial examples deviate from the data manifold more than from the distribution on that manifold.

{\bf Distributions in consecutive hidden layers (Hypothesis 2).} We examine which layer of the target network is the most discriminative for detecting adversarial examples. For this purpose, we plot the distribution of normal and adversarial samples in 2D space given by the reconstruction error and the latent norm for WAE in consecutive layers of the target network.

\begin{figure*}[h!] 
 \centering
 \centerline{\includegraphics[height=9
 cm]{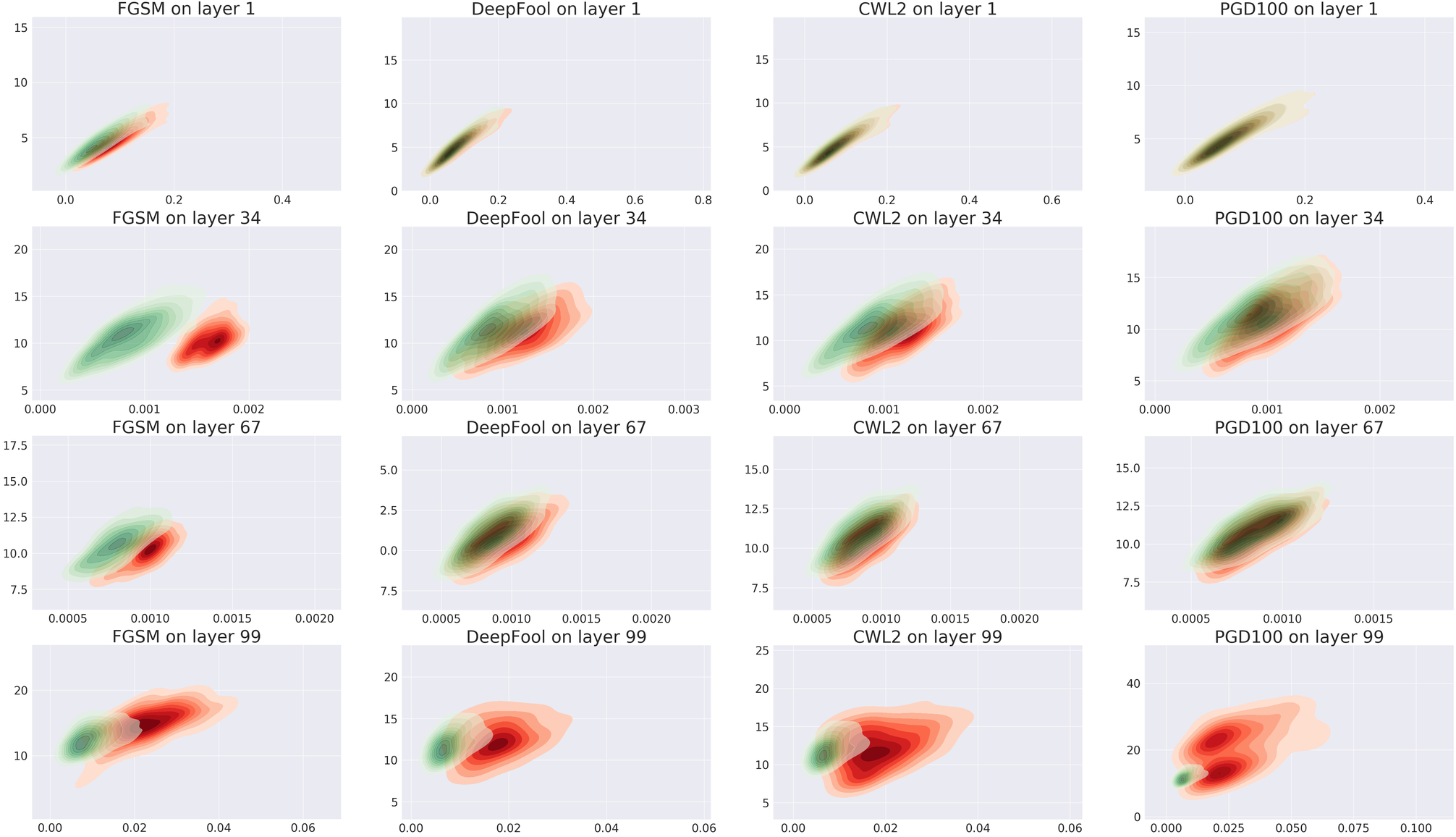}}
 \caption{The normal (green) and adversarial (red) examples visualized by the kernel density estimation in 2D space. In each sub-figure, the x-axis is the reconstruction error, the y-axis is the norm in the latent space. Each column of sub-figures refers to a different attack. These figures are generated for the SVHN dataset, DenseNet architecture. The adversarial examples are followed through 4 autoencoders, hence 4 rows of sub-figures. In most cases, the discrepancy between these distributions is most evident in the last layers. However, for simple attacks such as FGSM, adversarial examples can also be easily detected in earlier layers.}
\label{fig:adv_kde}
\end{figure*}

Figure \ref{fig:adv_kde} demonstrates that in nearly every setting, we can identify a layer where the distribution of normal and adversarial examples diverge. At first glance, deeper layers have the highest discriminative power. More careful analysis suggests, however, that for simple attacks, such as FGSM, initial layers also deliver substantial information for detection of abnormal samples (second row). For stronger attacks (DeepFool and CW) this discrepancy is not evident in the initial layers and the behavior in the last layers has to be investigated. For figures from other dataset-architecture combinations, we refer the reader to Supplementary Material, Section A.

To summarize this analysis in a quantitative way, we again analyze the importance of features using a random forest classifier. This time, we aggregate the importance of the reconstruction error and the latent norm for a given layer to get the total layer importance. Visualization of the layer importance (see Supplementary Material, Section C) confirms that for the FGSM attack, the most discriminative layer is the one after the first block, whereas the last two blocks are more important for other types of attacks. For simpler attacks, such as the FGSM, the adversarial examples quickly escape from the normal data manifold, hence earlier layers become important. More sophisticated attacks, such as DeepFool, CW and PGD, keep the adversarial example representations inside the normal data manifold for multiple layers, hence the final layer is the most discriminative for those attacks. 

\section{Experiments}
\label{sec:experiments}

\begin{table*}[]
\centering
\resizebox{1.0\textwidth}{!}{%
\begin{tabular}{ccc|ccccc|ccccc}
	\hline
	\multirow{2}{*}{Model} & \multirow{2}{*}{Dataset} & \multirow{2}{*}{Method} & \multicolumn{5}{c|}{Supervised setting} & \multicolumn{5}{c}{Unsupervised setting}\\ 
	& & & FGSM & BIM & DeepFool & CW & PGD & FGSM & BIM & DeepFool & CW & PGD \\ \hline
	\multirow{9}{*}{DenseNet} & \multirow{3}{*}{CIFAR-10} & Odds-testing & - & - & - & - & - & 45.23 & 69.01 & 58.30 & 61.29 & \textbf{97.93} \\ 
	& & Mahalanobis & 99.94 & 99.78 & 83.41 & 87.31 & 97.79 & - & - & - & - & - \\ 
	& & AE-layers (ours) & \textbf{100.00} & \textbf{99.99} & \textbf{91.36} & \textbf{97.75} & \textbf{99.61} & \textbf{78.38} & \textbf{97.51} & \textbf{65.31} & \textbf{68.15} & 94.20 \\ \cline{2-13} 
	& \multirow{3}{*}{CIFAR-100} & Odds-testing & - & - & - & - & - & 43.22 & 65.22 & 49.53 & 47.64 & \textbf{96.91}\\ 
	& & Mahalanobis & 99.86 & 99.17 & 77.57 & 87.05 & 79.24 & - & - & - & - & - \\ 
	& & AE-layers (ours) & \textbf{100.00} & \textbf{99.88} & \textbf{88.17} & \textbf{96.40} & 96.63 & \textbf{97.95} & \textbf{95.68} & \textbf{61.86} & \textbf{62.28} & 87.19 \\ \cline{2-13}
	& \multirow{3}{*}{SVHN} & Odds-testing & - & - & - & - & - & 56.14 & 71.11 & 67.81 & 70.71 & \textbf{99.25} \\ 
	& & Mahalanobis &  99.85 &  99.28 &  95.10 &  97.03 & 98.41 & - & - & - & - & - \\ 
	& & AE-layers (ours) & \textbf{99.98} & \textbf{99.75} & \textbf{97.26} & \textbf{97.80} & \textbf{99.43} & \textbf{96.50} & \textbf{94.07} & \textbf{83.80} & \textbf{84.81} & 93.70 \\ \hline
	\multirow{9}{*}{ResNet} & \multirow{3}{*}{CIFAR-10} & Odds-testing & - & - & - & - & - & 46.32 & 59.85 & 75.58 & 57.58 & \textbf{96.18} \\ 
	& & Mahalanobis & 99.94 & 99.57 & { \textbf{91.57}} & \textbf{95.84} & 89.81 & - & - & - & - & - \\ 
	& & AE-layers (ours) & \textbf{99.98} & \textbf{99.61} & 86.41 & 95.01 & \textbf{97.39} & \textbf{97.24} & \textbf{94.93} & \textbf{78.19} & \textbf{74.29} & 77.25 \\ \cline{2-13} 
	& \multirow{3}{*}{CIFAR-100} & Odds-testing & - & - & - & - & - & 38.26 & 43.52 & 61.13 & 44.74 & \textbf{93.73} \\ 
	& & Mahalanobis & 99.77 & 96.90 & \textbf{85.26} & 91.77 & 91.08 & - & - & - & - & - \\ 
	& & AE-layers (ours) & \textbf{100.00} & \textbf{99.52} & 77.98 & \textbf{96.41} & \textbf{98.12} & \textbf{95.59} & \textbf{80.23} & \textbf{71.06}	& \textbf{73.02} & 70.98 \\ \cline{2-13} 
	& \multirow{3}{*}{SVHN} & Odds-testing & - & - & - & - & - & 65.09 & 70.31 & 77.05 & 72.12 & \textbf{99.08}\\ 
	& & Mahalanobis & 99.62 & 97.15 & \textbf{95.73} & 92.15 & 92.24 & - & - & - & - & - \\ 
	& & AE-layers (ours) & \textbf{99.81} & \textbf{99.10} & 95.45 & \textbf{97.31} & 97.41 & \textbf{98.90} & \textbf{95.49} & \textbf{89.36}	& \textbf{90.02} & 83.62 \\ \hline
\end{tabular}
}
\caption{Comparison of AUROC (\%) scores. For the supervised setting, we use SVM as the final classifier and the entire latent vectors as its input features. For the unsupervised setting, we use Isolation Forest as the one-class classifier with the reconstruction errors and the latent norms as input features.}
\label{tab:results1}
\end{table*}

We test our method against FGSM \citep{goodfellow2014explaining}, BIM \citep{kurakin2016adversarial}, DeepFool \citep{moosavi2016deepfool}, CW \citep{carlini2017adversarial} and PGD \citep{madry2017towards} adversarial attacks on CIFAR-10 \citep{CIFAR10}, CIFAR-100 \citep{CIFAR100} and SVHN \citep{svhn} datasets for ResNet-34 \citep{he2016deep} and DenseNet-BC (L = 100, k = 12) \citep{huang2017densely} models. To reduce the computational time, we train WAE-MMD autoencoders on the selected layers of the target network. Specifically, we select the layers ending each block group in those architectures -- the same approach as in \citep{lee2018simple}. For simplicity, we use the same hyperparameters for each autoencoder and if the representation size allows, the same architecture, even between datasets and models. 

For the sake of a fair comparison, we use the test set creation procedure from \citep{lee2018simple}. In that setting, a noisy and an adversarial sample is created for each test set sample. The final test set consists of correctly classified clean and noisy examples (class 1) and incorrectly classified adversarial examples (class 2). In the supervised detection of adversarial examples, the final classifier is trained on 10\% of that test set and is evaluated on the remaining 90\%. For our methods (AE-layers), we use SVM as the final detection classifier with its hyperparameters being selected by 5-fold cross-validated grid search trained on the full AE latent vectors. In the unsupervised setting, we use the Isolation Forest one-class classifier, which is trained only on the training data without seeing any adversarial or noisy examples. To reduce the data dimension, we use a representation consisting of the reconstruction error and the $L_2$ norm taken from each autoencoder.

Table \ref{tab:results1} shows the results for our method compared to two state-of-the-art studies: Mahalanobis \citep{lee2018simple} (for the supervised case) and Odds-testing \citep{roth2019odds} (for the unsupervised case). In the supervised case, our method provides better results for almost all investigated cases. In the unsupervised case, our method is inferior only on the PGD-100 attack. While Odds-testing performs particularly well on that attack, it fails on other types of attacks, which were not tested in the original paper. We also highlight that our solution is more efficient than Odds-testing approach. Odds-testing requires multiple forward passes of the target network (256 in the original version) for each example to be tested, which is computationally expensive. In comparison, the inference in our method is cheap as the forward pass of several shallow (3-layer) encoders is faster than one forward pass of the target network.

In \citep{lee2018simple} authors do not provide a fully unsupervised solution and they consider only a partially supervised scenario. In Supplementary Material, Section F we report the results for such a setting. 

To compare the discriminative power of AE representations, we also run the supervised variant of our method trained on the reconstruction error and the latent norm. With the reduced representation the results are nearly the same. Precisely, the results drop only $0.01$ in terms of the mean AUROC score compared to the the analogical variant trained on full latent space of AE. (See Supplementary Material, Section E for detailed results.) This analysis gives a strong argument that the reconstruction error and the latent norm provides sufficient description of data manifold to detect adversarial examples. Moreover, it greatly accelerates training making our approach easy to use in various applications.

\section{Conclusion}
We presented a novel method for adversarial example detection that achieves state-of-the-art performance. It is based on two hypotheses that essentially outline the design of our method by pointing out that adversarial examples diverge from normal samples data manifold on different layers of the original network. We perform a thorough experimental analysis to confirm these hypotheses and then evaluate and compare our detector with two other methods in both supervised and unsupervised settings. The inference in our solution is fast and unlike Odds-testing it does not require multiple forward passes.

% Regarding future work, a few directions of research seem interesting to investigate. The first question is how the placement and the number of autoencoders affects the detection mechanisms. Another unexplored and challenging area of research is the unsupervised setting, where a detection model is built only with normal data.  

%%%%.%%%%%%%%%%%%%%%%%%%%%%%%%%%%%%%%%%%%%%%%%%%%%%%%%%%%%%%%%%%%%%%%%%%%%%%%%%%

\section*{Broader Impact}

A successful detection of adversarial examples is essential to establish trust and reliability of deep neural networks. It is hard to imagine a trustworthy self-driving car with a deep learning system, which can be easily fooled by an adversarial example. Or a classifier detecting cancer and giving a wrong prediction due to tiny perturbations which look like meaningless artifacts. Providing an additional mechanism, which detects adversarial examples helps to make the neural nets more applicable to real world problems. On the other hand, one must be very careful about guarantees given by such defenses. Typically, we aim at concrete attacks and scenarios, so more general claims may not be valid.

\bibliography{main}

\begin{thebibliography}{34}
\providecommand{\natexlab}[1]{#1}
\providecommand{\url}[1]{\texttt{#1}}
\expandafter\ifx\csname urlstyle\endcsname\relax
  \providecommand{\doi}[1]{doi: #1}\else
  \providecommand{\doi}{doi: \begingroup \urlstyle{rm}\Url}\fi

\bibitem[Amodei et~al.(2016)Amodei, Ananthanarayanan, Anubhai, Bai, Battenberg,
  Case, Casper, Catanzaro, Cheng, Chen, et~al.]{amodei2016deep}
Amodei, D., Ananthanarayanan, S., Anubhai, R., Bai, J., Battenberg, E., Case,
  C., Casper, J., Catanzaro, B., Cheng, Q., Chen, G., et~al.
\newblock Deep speech 2: End-to-end speech recognition in english and mandarin.
\newblock In \emph{International conference on machine learning}, pp.\
  173--182, 2016.

\bibitem[Carlini \& Wagner(2017)Carlini and Wagner]{carlini2017adversarial}
Carlini, N. and Wagner, D.
\newblock Adversarial examples are not easily detected: Bypassing ten detection
  methods.
\newblock 2017.

\bibitem[Carlini et~al.(2017)Carlini, Katz, Barrett, and
  Dill]{carlini2017provably}
Carlini, N., Katz, G., Barrett, C., and Dill, D.~L.
\newblock Provably minimally-distorted adversarial examples.
\newblock \emph{arXiv preprint arXiv:1709.10207}, 2017.

\bibitem[Ehlers(2017)]{ehlers2017formal}
Ehlers, R.
\newblock Formal verification of piece-wise linear feed-forward neural
  networks.
\newblock In \emph{International Symposium on Automated Technology for
  Verification and Analysis}, pp.\  269--286. Springer, 2017.

\bibitem[Fefferman et~al.(2013)Fefferman, Mitter, and
  Narayanan]{fefferman2013testing}
Fefferman, C., Mitter, S., and Narayanan, H.
\newblock Testing the manifold hypothesis, 2013.

\bibitem[Goodfellow et~al.(2014)Goodfellow, Shlens, and
  Szegedy]{goodfellow2014explaining}
Goodfellow, I.~J., Shlens, J., and Szegedy, C.
\newblock Explaining and harnessing adversarial examples.
\newblock \emph{arXiv preprint arXiv:1412.6572}, 2014.

\bibitem[Grosse et~al.(2017{\natexlab{a}})Grosse, Manoharan, Papernot, Backes,
  and McDaniel]{grosse2017statistical}
Grosse, K., Manoharan, P., Papernot, N., Backes, M., and McDaniel, P.
\newblock On the (statistical) detection of adversarial examples.
\newblock \emph{arXiv preprint arXiv:1702.06280}, 2017{\natexlab{a}}.

\bibitem[Grosse et~al.(2017{\natexlab{b}})Grosse, Papernot, Manoharan, Backes,
  and McDaniel]{grosse2017adversarial}
Grosse, K., Papernot, N., Manoharan, P., Backes, M., and McDaniel, P.
\newblock Adversarial examples for malware detection.
\newblock In \emph{European Symposium on Research in Computer Security}, pp.\
  62--79. Springer, 2017{\natexlab{b}}.

\bibitem[He et~al.(2016)He, Zhang, Ren, and Sun]{he2016deep}
He, K., Zhang, X., Ren, S., and Sun, J.
\newblock Deep residual learning for image recognition.
\newblock In \emph{Proceedings of the IEEE conference on computer vision and
  pattern recognition}, pp.\  770--778, 2016.

\bibitem[Hendrycks \& Gimpel(2016)Hendrycks and Gimpel]{hendrycks2016baseline}
Hendrycks, D. and Gimpel, K.
\newblock A baseline for detecting misclassified and out-of-distribution
  examples in neural networks.
\newblock \emph{arXiv preprint arXiv:1610.02136}, 2016.

\bibitem[Huang et~al.(2017)Huang, Liu, Van Der~Maaten, and
  Weinberger]{huang2017densely}
Huang, G., Liu, Z., Van Der~Maaten, L., and Weinberger, K.~Q.
\newblock Densely connected convolutional networks.
\newblock In \emph{Proceedings of the IEEE conference on computer vision and
  pattern recognition}, pp.\  4700--4708, 2017.

\bibitem[Kingma \& Welling(2014)Kingma and Welling]{VAE}
Kingma, D.~P. and Welling, M.
\newblock Auto-encoding variational bayes.
\newblock In \emph{2nd International Conference on Learning Representations,
  {ICLR} 2014, Banff, AB, Canada, April 14-16, 2014, Conference Track
  Proceedings}, 2014.
\newblock URL \url{http://arxiv.org/abs/1312.6114}.

\bibitem[Krizhevsky et~al.({\natexlab{a}})Krizhevsky, Nair, and
  Hinton]{CIFAR10}
Krizhevsky, A., Nair, V., and Hinton, G.
\newblock Cifar-10 (canadian institute for advanced research).
\newblock {\natexlab{a}}.
\newblock URL \url{http://www.cs.toronto.edu/~kriz/cifar.html}.

\bibitem[Krizhevsky et~al.({\natexlab{b}})Krizhevsky, Nair, and
  Hinton]{CIFAR100}
Krizhevsky, A., Nair, V., and Hinton, G.
\newblock Cifar-100 (canadian institute for advanced research).
\newblock {\natexlab{b}}.
\newblock URL \url{http://www.cs.toronto.edu/~kriz/cifar.html}.

\bibitem[Kurakin et~al.(2016)Kurakin, Goodfellow, and
  Bengio]{kurakin2016adversarial}
Kurakin, A., Goodfellow, I., and Bengio, S.
\newblock Adversarial examples in the physical world.
\newblock \emph{arXiv preprint arXiv:1607.02533}, 2016.

\bibitem[Lee et~al.(2018)Lee, Lee, Lee, and Shin]{lee2018simple}
Lee, K., Lee, K., Lee, H., and Shin, J.
\newblock A simple unified framework for detecting out-of-distribution samples
  and adversarial attacks.
\newblock In \emph{Advances in Neural Information Processing Systems}, pp.\
  7167--7177, 2018.

\bibitem[Lomuscio \& Maganti(2017)Lomuscio and Maganti]{lomuscio2017approach}
Lomuscio, A. and Maganti, L.
\newblock An approach to reachability analysis for feed-forward relu neural
  networks.
\newblock \emph{arXiv preprint arXiv:1706.07351}, 2017.

\bibitem[Madry et~al.(2017)Madry, Makelov, Schmidt, Tsipras, and
  Vladu]{madry2017towards}
Madry, A., Makelov, A., Schmidt, L., Tsipras, D., and Vladu, A.
\newblock Towards deep learning models resistant to adversarial attacks.
\newblock \emph{arXiv preprint arXiv:1706.06083}, 2017.

\bibitem[Moosavi-Dezfooli et~al.(2016)Moosavi-Dezfooli, Fawzi, and
  Frossard]{moosavi2016deepfool}
Moosavi-Dezfooli, S.-M., Fawzi, A., and Frossard, P.
\newblock Deepfool: a simple and accurate method to fool deep neural networks.
\newblock In \emph{Proceedings of the IEEE conference on computer vision and
  pattern recognition}, pp.\  2574--2582, 2016.

\bibitem[Morawiecki et~al.(2019)Morawiecki, Spurek, {\'S}mieja, and
  Tabor]{morawiecki2019fast}
Morawiecki, P., Spurek, P., {\'S}mieja, M., and Tabor, J.
\newblock Fast and stable interval bounds propagation for training verifiably
  robust models.
\newblock \emph{arXiv preprint arXiv:1906.00628}, 2019.

\bibitem[Netzer et~al.(2011)Netzer, Wang, Coates, Bissacco, Wu, and Ng]{svhn}
Netzer, Y., Wang, T., Coates, A., Bissacco, A., Wu, B., and Ng, A.~Y.
\newblock Reading digits in natural images with unsupervised feature learning.
\newblock In \emph{NIPS Workshop on Deep Learning and Unsupervised Feature
  Learning 2011}, 2011.
\newblock URL
  \url{http://ufldl.stanford.edu/housenumbers/nips2011_housenumbers.pdf}.

\bibitem[Papernot \& McDaniel(2018)Papernot and McDaniel]{deep_knn}
Papernot, N. and McDaniel, P.~D.
\newblock Deep k-nearest neighbors: Towards confident, interpretable and robust
  deep learning.
\newblock \emph{CoRR}, abs/1803.04765, 2018.
\newblock URL \url{http://arxiv.org/abs/1803.04765}.

\bibitem[Pidhorskyi et~al.(2018)Pidhorskyi, Almohsen, and
  Doretto]{pidhorskyi2018generative}
Pidhorskyi, S., Almohsen, R., and Doretto, G.
\newblock Generative probabilistic novelty detection with adversarial
  autoencoders.
\newblock In \emph{Advances in neural information processing systems}, pp.\
  6822--6833, 2018.

\bibitem[Roth et~al.(2019)Roth, Kilcher, and Hofmann]{roth2019odds}
Roth, K., Kilcher, Y., and Hofmann, T.
\newblock The odds are odd: A statistical test for detecting adversarial
  examples.
\newblock \emph{arXiv preprint arXiv:1902.04818}, 2019.

\bibitem[Samangouei et~al.(2018)Samangouei, Kabkab, and
  Chellappa]{samangouei2018defense}
Samangouei, P., Kabkab, M., and Chellappa, R.
\newblock Defense-gan: Protecting classifiers against adversarial attacks using
  generative models.
\newblock \emph{arXiv preprint arXiv:1805.06605}, 2018.

\bibitem[Sitawarin et~al.(2018)Sitawarin, Bhagoji, Mosenia, Chiang, and
  Mittal]{sitawarin2018darts}
Sitawarin, C., Bhagoji, A.~N., Mosenia, A., Chiang, M., and Mittal, P.
\newblock Darts: Deceiving autonomous cars with toxic signs.
\newblock \emph{arXiv preprint arXiv:1802.06430}, 2018.

\bibitem[Szegedy et~al.(2013)Szegedy, Zaremba, Sutskever, Bruna, Erhan,
  Goodfellow, and Fergus]{szegedy2013intriguing}
Szegedy, C., Zaremba, W., Sutskever, I., Bruna, J., Erhan, D., Goodfellow, I.,
  and Fergus, R.
\newblock Intriguing properties of neural networks.
\newblock \emph{arXiv preprint arXiv:1312.6199}, 2013.

\bibitem[Tabor et~al.(2018)Tabor, Knop, Spurek, Podolak, Mazur, and
  Jastrz{\k{e}}bski]{tabor2018cramer}
Tabor, J., Knop, S., Spurek, P., Podolak, I., Mazur, M., and Jastrz{\k{e}}bski,
  S.
\newblock Cramer-wold autoencoder.
\newblock \emph{arXiv preprint arXiv:1805.09235}, 2018.

\bibitem[Tolstikhin et~al.(2018)Tolstikhin, Bousquet, Gelly, and
  Sch{\"{o}}lkopf]{WAE}
Tolstikhin, I.~O., Bousquet, O., Gelly, S., and Sch{\"{o}}lkopf, B.
\newblock Wasserstein auto-encoders.
\newblock In \emph{6th International Conference on Learning Representations,
  {ICLR} 2018, Vancouver, BC, Canada, April 30 - May 3, 2018, Conference Track
  Proceedings}, 2018.

\bibitem[Xiao et~al.(2018)Xiao, Tjeng, Shafiullah, and Madry]{xiao2018training}
Xiao, K.~Y., Tjeng, V., Shafiullah, N.~M., and Madry, A.
\newblock Training for faster adversarial robustness verification via inducing
  relu stability.
\newblock \emph{arXiv preprint arXiv:1809.03008}, 2018.

\bibitem[Xu et~al.(2017)Xu, Evans, and Qi]{xu2017feature}
Xu, W., Evans, D., and Qi, Y.
\newblock Feature squeezing: Detecting adversarial examples in deep neural
  networks.
\newblock \emph{arXiv preprint arXiv:1704.01155}, 2017.

\bibitem[Yang et~al.(2019)Yang, Chen, Hsieh, Wang, and Jordan]{yang2019ml}
Yang, P., Chen, J., Hsieh, C.-J., Wang, J.-L., and Jordan, M.~I.
\newblock Ml-loo: Detecting adversarial examples with feature attribution.
\newblock \emph{arXiv preprint arXiv:1906.03499}, 2019.

\bibitem[Zhang et~al.(2018)Zhang, Weng, Chen, Hsieh, and
  Daniel]{zhang2018efficient}
Zhang, H., Weng, T.-W., Chen, P.-Y., Hsieh, C.-J., and Daniel, L.
\newblock Efficient neural network robustness certification with general
  activation functions.
\newblock In \emph{Advances in Neural Information Processing Systems}, pp.\
  4939--4948, 2018.

\bibitem[Zhao et~al.(2019)Zhao, Zheng, Xu, and Wu]{zhao2019object}
Zhao, Z.-Q., Zheng, P., Xu, S.-t., and Wu, X.
\newblock Object detection with deep learning: A review.
\newblock \emph{IEEE transactions on neural networks and learning systems},
  30\penalty0 (11):\penalty0 3212--3232, 2019.

\end{thebibliography}
\bibliographystyle{icml2020}

\clearpage 
\appendix

\section {Distribution of Normal and Adversarial Examples Representation}

Below we provide the figures, which show normal and adversarial examples representations (their distributions). The figures are given for all six dataset-architecture combinations we explore. The normal (green) and adversarial (red) examples are visualized by the kernel density estimation in the 2D space. In each sub-figure, the x-axis is the reconstruction error, the y-axis is the $L_2$ norm in the latent space. Each column of sub-figures refers to a different attack (FGSM, BIM, Deepfool, Carlini-Wagner and PGD). The adversarial examples are followed through subsequent layers, each corresponding to a different row of sub-figures. In most investigated cases, the discrepancy between these distributions is most evident in last layers. However, for simpler attacks, such as FGSM, the adversarial examples can also be easily detected in earlier layers.

\begin{figure*}[!ht]
	\centering
	\centerline{\includegraphics[width=1.33\textwidth]{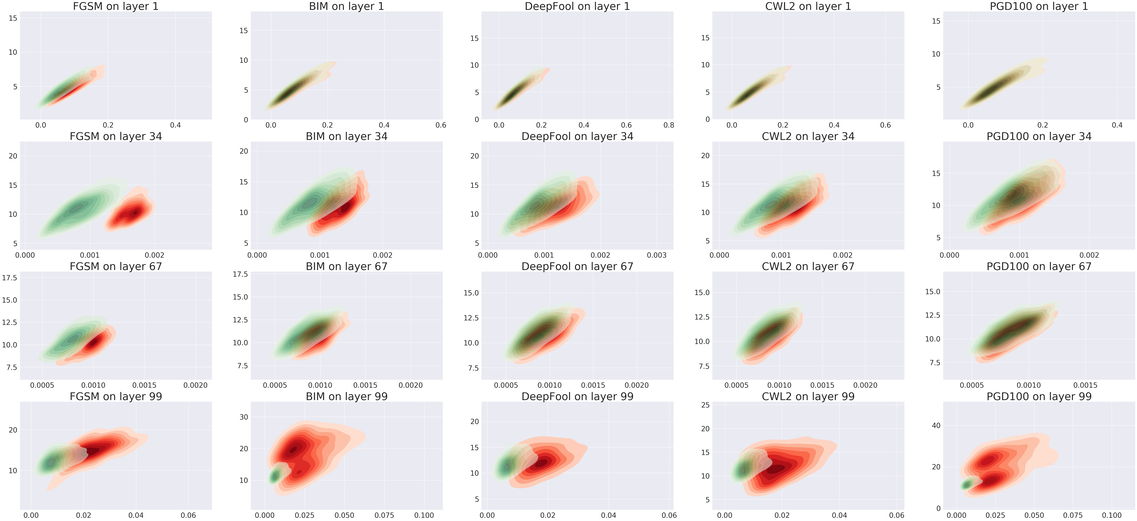}}
	\caption{SVHN dataset, DenseNet model}
\end{figure*} 

\begin{figure*}[!ht] 
	\centering
	\centerline{\includegraphics[width=1.33\textwidth]{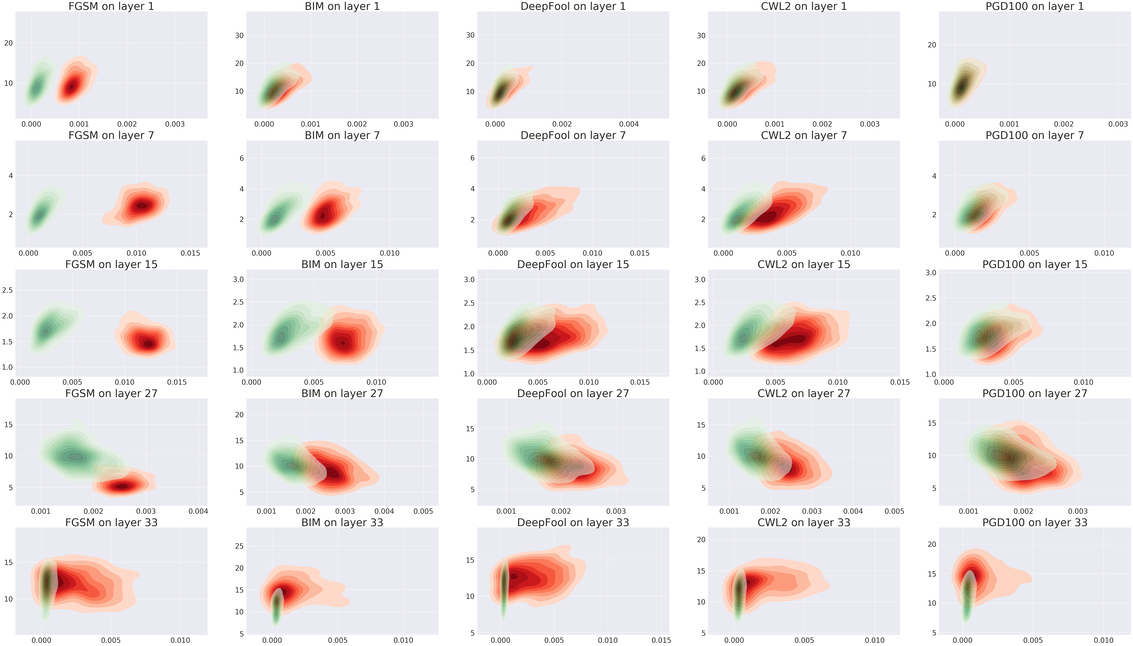}}
	\caption{SVHN dataset, ResNet model}
\end{figure*} 

\begin{figure*}[!ht] 
	\centering
	\centerline{\includegraphics[width=1.33\textwidth]{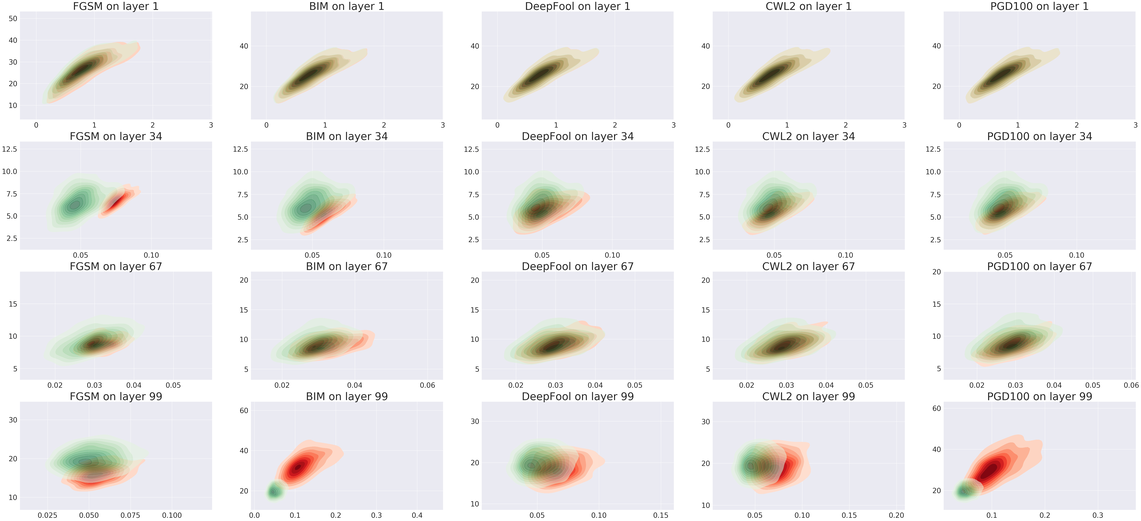}}
	\caption{CIFAR-10 dataset, DenseNet model}
\end{figure*} 

\begin{figure*}[!ht] 
	\centering
	\centerline{\includegraphics[width=1.33\textwidth]{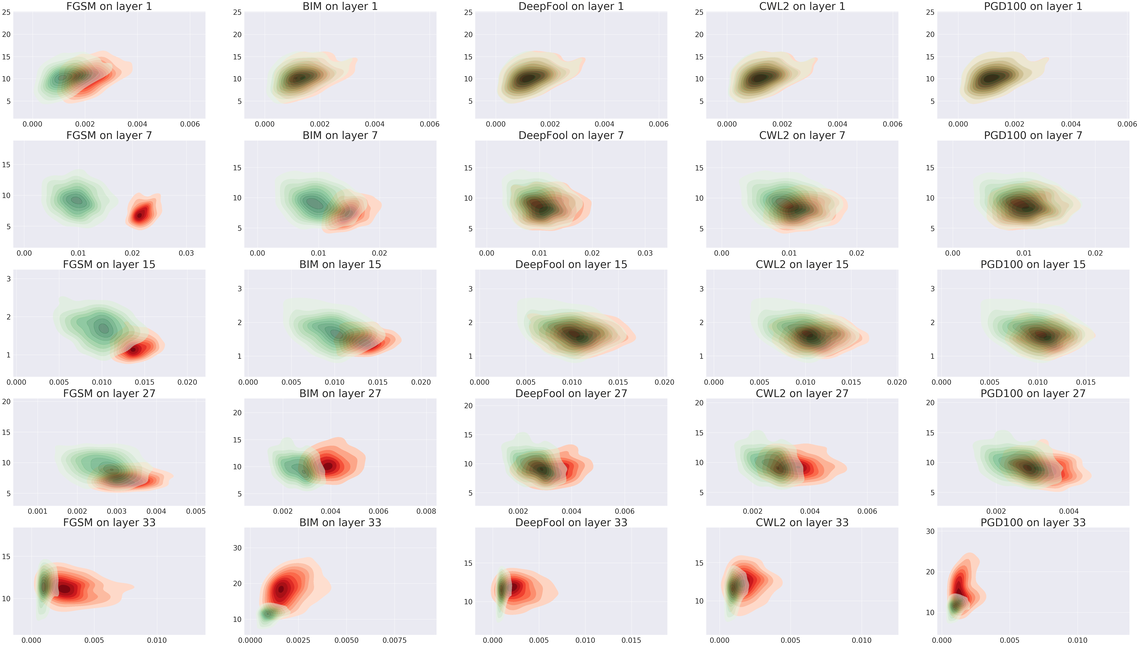}}
	\caption{CIFAR-10 dataset, ResNet model}
\end{figure*}

\begin{figure*}[!ht] 
	\centering
	\centerline{\includegraphics[width=1.33\textwidth]{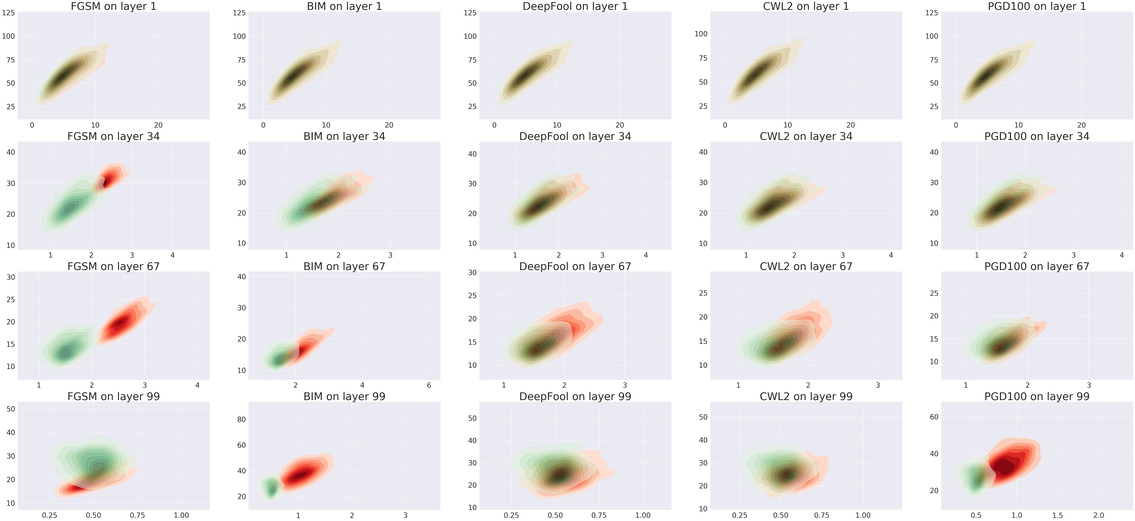}}
	\caption{CIFAR-100 dataset, DenseNet model}
\end{figure*} 

\begin{figure*}[!t] 
	\centering
	\centerline{\includegraphics[width=1.33\textwidth]{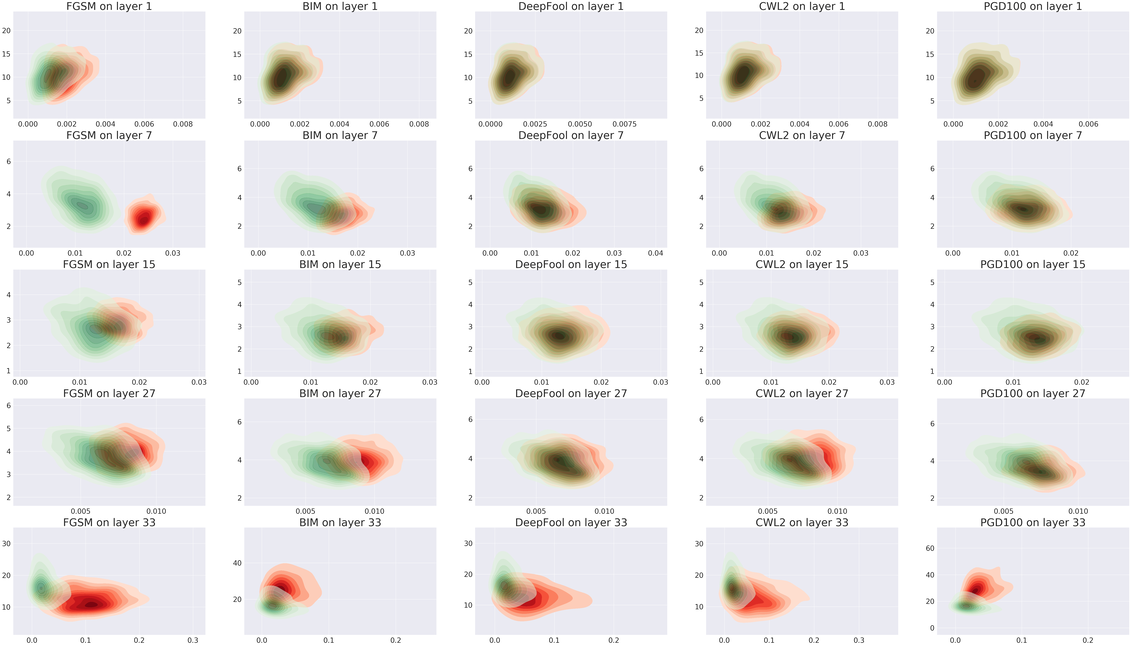}}
	\caption{CIFAR-100 dataset, ResNet model}
\end{figure*}

\clearpage

\section{Significance of Adversarial Perturbation Magnitude}

Below we provide the figures visualizing an impact of the perturbation magnitude on the position of an example in 2D space (reconstruction error in the x-axis and latent $L_2$ norm in the y-axis). The figures are given for all six dataset-architecture combinations we investigate. 

The starting point is a single normal example without any perturbations (visualized as a light yellow dot). Then, we linearly increase the perturbation until it reaches the value of $2\epsilon$ (dark red).  As a point of reference, we provide the kernel density estimation (green) of normal examples from the test set. Each column of sub-figures refers to a different attack (FGMS, BIM, DeepFool, PGD) and each row of sub-figures refers to a different layer. For most settings, we observe that adversarial examples significantly move away from the data manifold as the perturbation magnitude increases (movement in x-axis).

\begin{figure*}[!ht] 
	\centering
	\centerline{\includegraphics[width=1.1\textwidth]{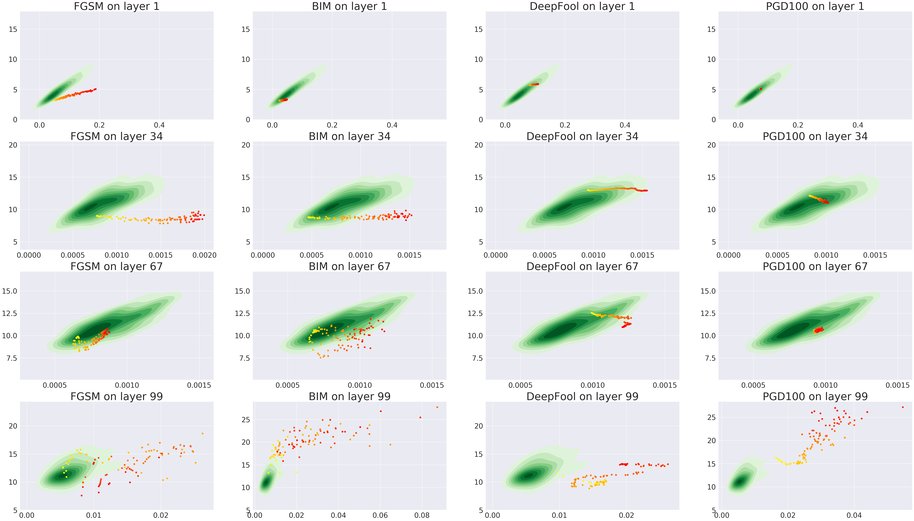}}
	\caption{SVHN dataset, DenseNet model}
\end{figure*} 

\begin{figure*}[!ht] 
	\centering
	\centerline{\includegraphics[width=1.1\textwidth]{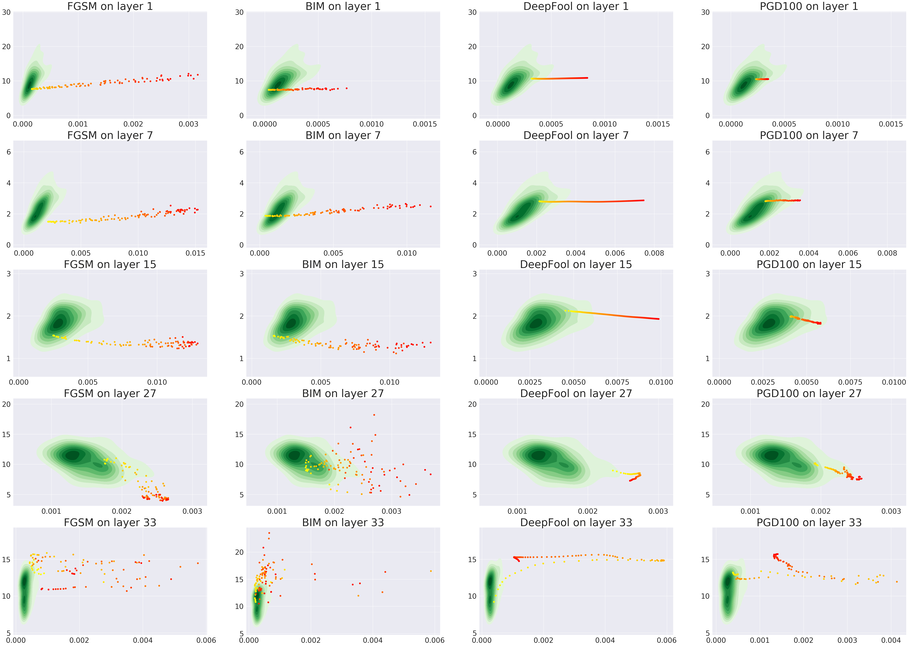}} 
	\caption{SVHN dataset, ResNet model}
\end{figure*} 

\begin{figure*}[!ht] 
	\centering
	\centerline{\includegraphics[width=1.1\textwidth]{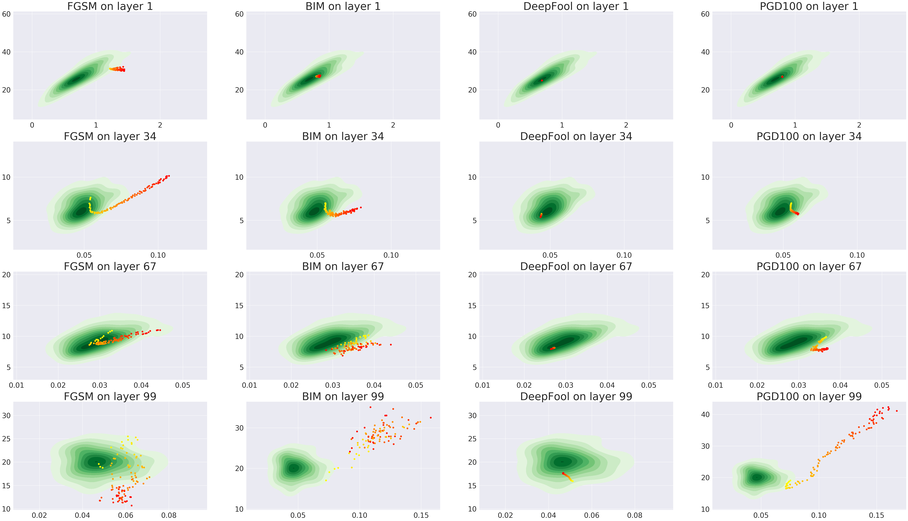}}
	\caption{CIFAR-10 dataset, DenseNet model}
\end{figure*} 

\begin{figure*}[!ht] 
	\centering
	\centerline{\includegraphics[width=1.1\textwidth]{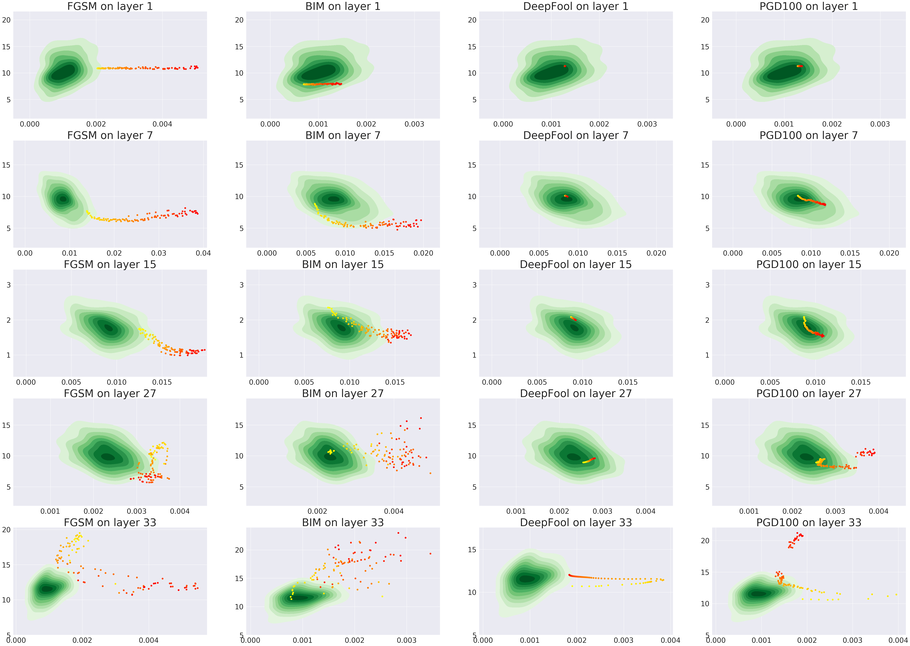}}
	\caption{CIFAR-10 dataset, ResNet model}
\end{figure*}

\begin{figure*}[!ht] 
	\centering
	\centerline{\includegraphics[width=1.1\textwidth]{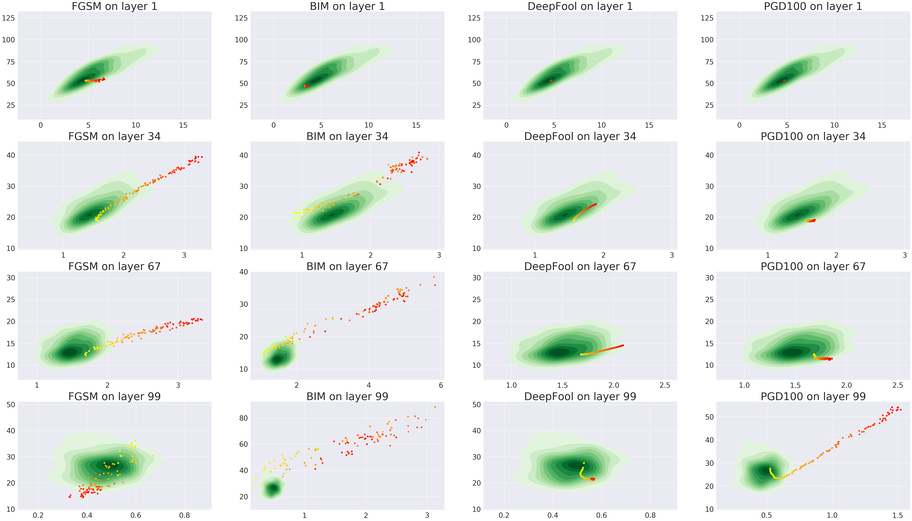}}
	\caption{CIFAR-100 dataset, DenseNet model}
\end{figure*} 

\begin{figure*}[t!] 
	\centering
	\centerline{\includegraphics[width=1.1\textwidth]{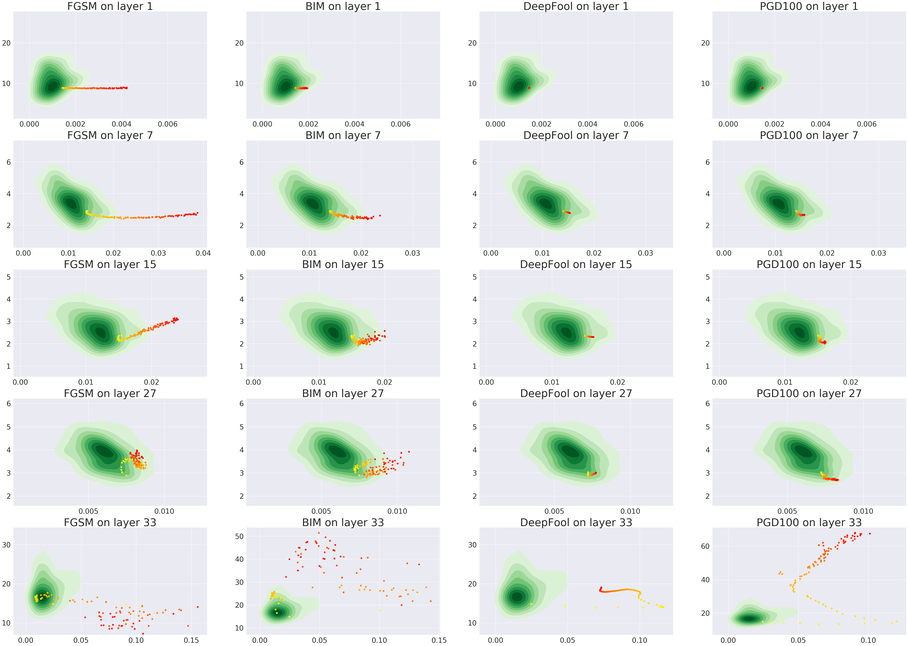}}
	\caption{CIFAR-100 dataset, ResNet model}
\end{figure*}

\clearpage
\section{Layer importance}

In this section we provide the layer importances calculated for the model with the random forest as the final classifier. 
Feature importances ($L_2$ norm in the latent space and the reconstruction error) are summed for a given layer. The final layers have the highest discriminative power except for the case of the simplest attack (FGSM), which can be easily detected using the activations from the second block of hidden layers. These findings are consistent with the figures from Section A.

\begin{figure*}[h!]
	\begin{minipage}{0.5\textwidth}
		\centering
		\hspace*{-1.5cm}
		\includegraphics[width=\linewidth]{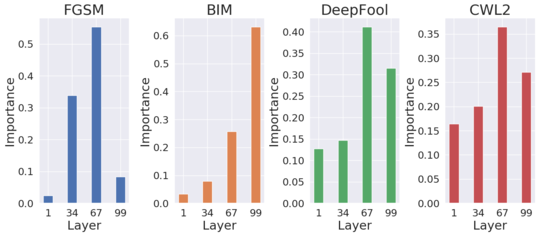}
		\hspace*{-1.5cm}
		\caption{CIFAR-100 dataset, DenseNet model}
	\end{minipage}\hfill
	\begin{minipage}{0.5\textwidth}
		\centering
		\includegraphics[width=\linewidth]{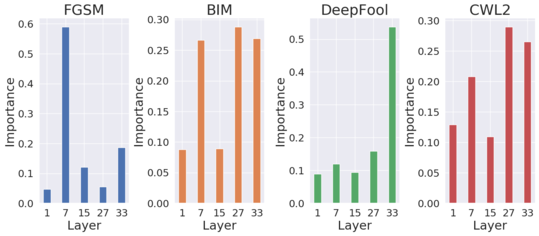}
		\caption{CIFAR-100 dataset, ResNet model}
	\end{minipage}
\end{figure*}

\begin{figure*}[h!]
	\begin{minipage}{0.5\textwidth}
		\centering
		\hspace*{-1.5cm}
		\includegraphics[width=\linewidth]{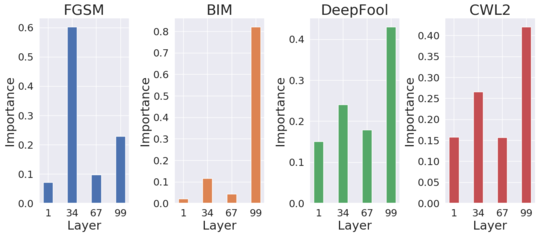}
		\hspace*{-1.5cm}
		\caption{CIFAR-10 dataset, DenseNet model}
	\end{minipage}\hfill
	\begin{minipage}{0.5\textwidth}
		\centering
		\includegraphics[width=\linewidth]{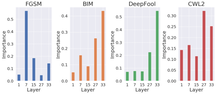}
		\caption{CIFAR-10 dataset, ResNet model}
	\end{minipage}
\end{figure*}

\begin{figure*}[h!]
	\begin{minipage}{0.5\textwidth}
		\centering
		\hspace*{-1.5cm}
		\includegraphics[width=\linewidth]{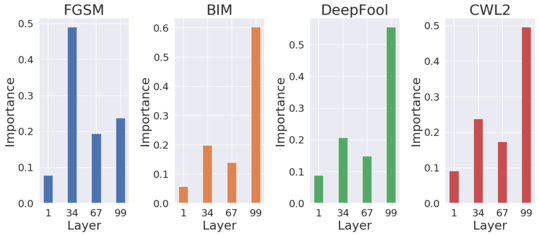}
		\hspace*{-1.5cm}
		\caption{SVHN dataset, DenseNet model}
	\end{minipage}\hfill
	\begin{minipage}{0.5\textwidth}
		\centering
		\includegraphics[width=\linewidth]{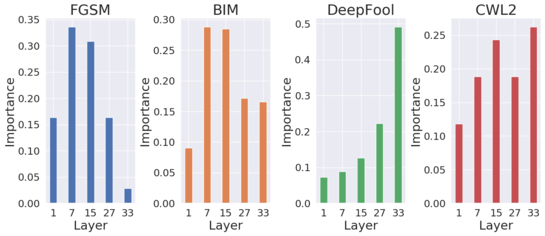}
		\caption{SVHN dataset, ResNet model}
	\end{minipage}
\end{figure*}

\clearpage
\section{Feature importance}

We also investigate the feature importance for all dataset-architecture combinations with the random forest model. 
Features ($L_2$ norm in the latent space and the reconstruction error) are aggregated over layers. Clearly, the reconstruction error, which indicates the distance of a given sample from the manifold of true data, is more discriminative for detecting adversarial examples than the latent norm, which describes the movement within the manifold. This analysis supports Hypothesis 1 stated in the main body of the paper.

\begin{figure}[h!]
	\begin{minipage}{0.5\textwidth}
		\centering
		\hspace*{-1.5cm}
		\includegraphics[width=\linewidth]{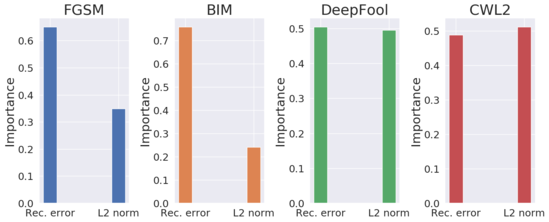}
		\hspace*{-1.5cm}
		\caption{CIFAR-100 dataset, DenseNet model}
	\end{minipage}\hfill
	\begin{minipage}{0.5\textwidth}
		\centering
		\includegraphics[width=\linewidth]{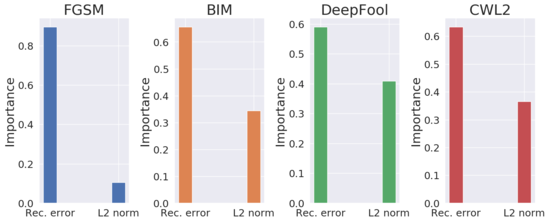}
		\caption{CIFAR-100 dataset, ResNet model}
	\end{minipage}
\end{figure}

\begin{figure}[h!]
	\begin{minipage}{0.5\textwidth}
		\centering
		\hspace*{-1.5cm}
		\includegraphics[width=\linewidth]{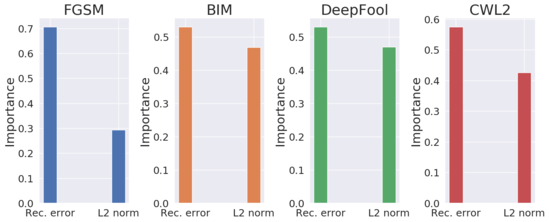}
		\hspace*{-1.5cm}
		\caption{CIFAR-10 dataset, DenseNet model}
	\end{minipage}\hfill
	\begin{minipage}{0.5\textwidth}
		\centering
		\includegraphics[width=\linewidth]{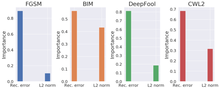}
		\caption{CIFAR-10 dataset, ResNet model}
	\end{minipage}
\end{figure}

\begin{figure}[h!]
	\begin{minipage}{0.5\textwidth}
		\centering
		\hspace*{-1.5cm}
		\includegraphics[width=\linewidth]{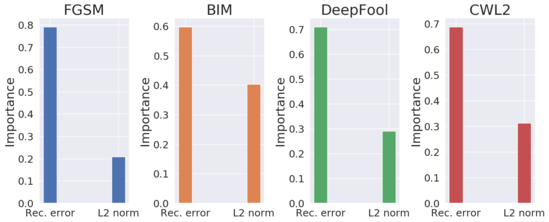}
		\hspace*{-1.5cm}
		\caption{SVHN dataset, DenseNet model}
	\end{minipage}\hfill
	\begin{minipage}{0.5\textwidth}
		\centering
		\includegraphics[width=\linewidth]{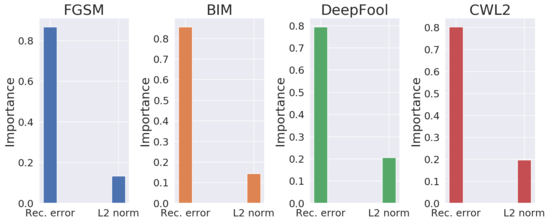}
		\caption{SVHN dataset, ResNet model} 
	\end{minipage}
\end{figure}

\clearpage

\section{Comparison of different representations}

The final classifier can be trained on different representations taken from the auto-encoders.
We compare representation using the entire latent vector (full), with ones based on either reconstruction error only, or the latent norm only or both features. The results reported for the supervised (Table \ref{tab:repres2})  confirm that: (1) reconstruction error is more discriminative than latent norm (2) full latent representation is only slightly better than using two AE features. For the unsupervised case (Table \ref{tab:repres1}) we do not consider the full latent vectors as the one-class methods we examine perform poorly when the feature space is large. In the unsupervised setting the reconstruction error is also the most important feature.

\begin{table}[h!]
	\centering
	\resizebox{0.8\textwidth}{!}{%
		\begin{tabular}{ccc|ccccc}
			\hline
			Model & Dataset & & FGSM & BIM & DeepFool & CW & PGD \\ \hline
			\multirow{12}{*}{DenseNet} & \multirow{4}{*}{CIFAR-10} & full & \textbf{100.00} & \textbf{99.99} & \textbf{91.36} & \textbf{97.75} & \textbf{99.61} \\ 
			& & both & 99.03 & 99.96 & 88.26 & 94.34 & 99.38 \\ 
			& & rec. error & 98.20 & 99.23 & 78.81 & 84.95 & 97.66 \\ 
			& & lat. norm & 88.32 & 99.52 & 78.58 & 86.08 & 97.75 \\ \cline{2-8} 
			& \multirow{4}{*}{CIFAR-100} & full & \textbf{100.00} & \textbf{99.88} & \textbf{88.17} & \textbf{96.40} & 96.63 \\
			& & both & 99.94 & 99.60 & 81.63 & 88.25 & \textbf{97.26} \\
			& & rec.error & 99.71 & 99.50 & 73.32 & 78.97 & 96.92 \\
			& & lat. norm & 99.80 & 90.71 & 77.26 & 82.35 & 83.44 \\ \cline{2-8}
			& \multirow{4}{*}{SVHN} & full & \textbf{99.98} & \textbf{99.75} & \textbf{97.26} & \textbf{97.80} & \textbf{99.43} \\
			& &  both & 99.93 & 98.77 & 95.24 & 96.98 & 97.20 \\
			& & rec.error & 99.40 & 96.12 & 91.75 & 92.54 & 95.81 \\ 
			& & lat. norm & 79.90 & 85.12 & 57.22 & 65.43 & 87.08 \\ \hline
			\multirow{12}{*}{ResNet} & \multirow{4}{*}{CIFAR-10} & full & 99.98 & 99.61 & 86.41 & \textbf{95.01} & \textbf{97.39} \\
			& & both & \textbf{100.0} & \textbf{99.91} & \textbf{91.37} & 93.95 & 94.03 \\
			& & rec.error & 99.99 & 99.14 & 91.05 & 91.55 & 83.07 \\
			& & lat. norm & 92.87 & 97.95 & 64.41 & 78.11 & 90.66 \\ \cline{2-8}
			& \multirow{4}{*}{CIFAR-100} & full & \textbf{100.00} & \textbf{99.52} & 77.98 & \textbf{96.41} & \textbf{98.12} \\ 
			& & both & 99.98 & 98.52 & \textbf{85.00} & 95.08 & 96.62 \\
			& & rec.error & 99.98 & 98.52 & 81.02 & 92.05 & 78.59 \\
			& & lat. norm & 98.11 & 95.90 & 78.29 & 87.84 & 94.81 \\ \cline{2-8}
			& \multirow{4}{*}{SVHN} & full & 99.81 & 99.10 & 95.45 & 97.31 & \textbf{97.41} \\ 
			& & both & \textbf{99.95} & \textbf{99.56} & \textbf{96.33} & \textbf{98.01} & 93.86 \\
			& & rec.error & 99.92 & 99.29 & 95.78 & 97.48 & 88.30 \\
			& & lat. norm & 98.38 & 93.30 & 78.99 & 85.95 & 86.49 \\ \hline
		\end{tabular}%
	}
	\caption{Comparison of four variants of AE representations for supervised learning: (1) entire latent vector, (2) reconstruction error supplied with latent norm, (3) reconstruction error only (4) latent norm only.
	}
	\label{tab:repres2}
\end{table}

\begin{table}[h!]
	\centering
	\resizebox{0.8\textwidth}{!}{%
		\begin{tabular}{ccc|ccccc}
			\hline
			Model & Dataset & & FGSM & BIM & DeepFool & CW & PGD \\ \hline
			\multirow{9}{*}{DenseNet} & \multirow{3}{*}{CIFAR-10} & both & 78.38 & \textbf{97.51} & \textbf{65.31} & 68.15 & \textbf{94.20} \\
			& & rec. error & \textbf{82.87} & 95.79 & 65.28 & \textbf{70.35} & 91.88 \\
			& & lat. norm & 63.27 & 95.82 & 59.05 & 59.37 & 90.47 \\ \cline{2-8} 
			& \multirow{3}{*}{CIFAR-100} & both & \textbf{97.95} & 95.68 & \textbf{61.86} & \textbf{62.28} & 87.19 \\
			& & rec. error & 96.98 & \textbf{96.92} & 59.74 & 59.81 & \textbf{91.30} \\
			& & lat. norm & 95.05 & 77.46 & 59.14 & 59.10 & 66.64 \\ \cline{2-8} 
			& \multirow{3}{*}{SVHN} & both & 96.50 & \textbf{94.07} & 83.80 & 84.81 & \textbf{93.70} \\
			& & rec. error & \textbf{96.66} & 91.43 & \textbf{85.84} & \textbf{84.82} & 91.11 \\
			& & lat. norm & 69.88 & 82.19 & 56.61 & 61.81 & 84.58 \\ \hline 
			\multirow{9}{*}{ResNet} & \multirow{3}{*}{CIFAR-10} & both & 97.24 & \textbf{94.93} & 78.19 & 74.29 & \textbf{77.25} \\
			& & rec. error & \textbf{97.42} & 91.19 & \textbf{83.21} & \textbf{78.52} & 69.79 \\
			& & lat. norm & 70.91 & 90.34 & 46.24 & 49.82 & 75.34 \\ \cline{2-8} 
			& \multirow{3}{*}{CIFAR-100} & both & 95.59 & \textbf{80.23} & \textbf{71.06} & 73.02 & 70.98 \\
			& & rec. error & \textbf{96.53} & 79.77 & 70.70 & \textbf{73.76} & 58.00 \\
			& & lat. norm & 71.94 & 69.45 & 61.18 & 60.62 & \textbf{76.67} \\ \cline{2-8} 
			& \multirow{3}{*}{SVHN} & both & 98.90 & 95.49 & 89.36 & 90.02 & \textbf{83.62} \\
			& & rec. error & \textbf{99.09} & \textbf{95.78} & \textbf{90.84} & \textbf{91.76} & 80.34 \\
			& & lat. norm & 88.30 & 77.95 & 60.71 & 65.48 & 72.67 \\ \hline
		\end{tabular}%
	}
	\caption{Performance comparison of the one-class classifier (Isolation Forest) trained on: (1) reconstruction error and latent norm, (2) reconstruction error only, (3) latent norm only.
		%for the supervised (linear regression) and unsupervised (Isolation Forest \citep{liu2008isolation}) classifiers with the reconstruction errors and the latent norms as input features.
	}
	\label{tab:repres1}
\end{table}

\section{Detection in the partially supervised scenario} \label{app:detection}

We examine how the detector trained on the FGSM attack generalizes to other types of attacks. Table \ref{tab:resultsApp} shows that the performance of our method is comparable to the Mahalanobis detector (Lee et al., 2018). We argue that there might be a trade-off between performance on a fully supervised setting (where our method gets 100\% on some cases) and a generalization ability to other attacks. 

\begin{table*}[h!]
	\centering
	\resizebox{0.8\textwidth}{!}{%
		\begin{tabular}{ccc|cccc}
			\hline
			Model & Dataset & Method & FGSM (seen) & BIM & DeepFool & CW  \\ \hline
			\multirow{6}{*}{DenseNet} & \multirow{2}{*}{CIFAR-10} &  Mahalanobis & 99.94 & \textbf{99.51} & 83.42 & 87.95 \\ 
			& & AE-layers (ours) & \textbf{100.00} & 95.25 & \textbf{84.59} & \textbf{92.44}  \\ \cline{2-7 } 
			& \multirow{2}{*}{CIFAR-100} & Mahalanobis & 99.86 & 98.27 & 75.63 & 86.20  \\ 
			& & AE-layers (ours) & \textbf{100.00} & \textbf{98.54} & \textbf{82.96} & \textbf{93.75} \\ \cline{2-7}
			& \multirow{2}{*}{SVHN} & Mahalanobis & 99.85 & \textbf{99.12} & \textbf{93.47} & \textbf{96.95}\\ 
			& & AE-layers (ours) & \textbf{99.98} & 96.94 & 87.55 & 93.45  \\ \hline
			\multirow{6}{*}{ResNet} & \multirow{2}{*}{CIFAR-10} & Mahalanobis & 99.94 & \textbf{98.91} & \textbf{78.06} & \textbf{93.90}  \\ 
			& & AE-layers (ours) & \textbf{99.98} & 91.53 & 70.82 & 88.19  \\ \cline{2-7} 
			& \multirow{2}{*}{CIFAR-100} & Mahalanobis & 99.77 & \textbf{96.38} & \textbf{81.95} & 90.96 \\ 
			& & AE-layers (ours) & \textbf{100.00} & 94.02 & 73.53 & \textbf{93.82} \\ \cline{2-7} 
			& \multirow{2}{*}{SVHN} & Mahalanobis & 99.62 & \textbf{95.39} & 72.20 & 86.73 \\
			& & AE-layers (ours) & \textbf{99.81} & 92.46 & \textbf{75.66} & \textbf{86.99}  \\ \hline
		\end{tabular}
	}
	\caption{Comparison of AUROC (\%) scores. The classifier is trained on the FGSM attack and tested against other attacks. For our method, we use SVM as the final classifier and the entire latent vectors as its input features.}
	\label{tab:resultsApp}
\end{table*}

\section{Attack strength vs. detection performance}
\begin{figure*}[!ht] 
	\centering
	\centerline{\includegraphics[width=0.7\textwidth]{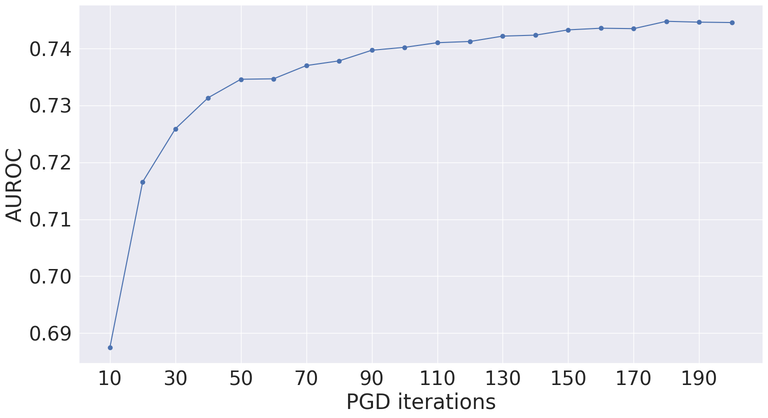}}
	\caption{PGD iterations vs. detection performance}
	\label{fig:iteration_impact}
\end{figure*}

To investigate how the number of iterations in the PGD attack affects the detection performance, we generate adversarial examples for the entire test set for multiple iteration count values. We then measure the detection performance of our unsupervised Isolation Forest final classifier for each iteration value. We perform this experiment on CIFAR-10 and the ResNet architecture and present the results in Figure \ref{fig:iteration_impact}. Interestingly, the stronger attack (more iterations), the better detection performance. We observe similar phenomenon with the Odds-testing method (Roth et al., 2019) when `weaker' attacks turn out to be much more challenging for that method (See Table 1 in the main body of the paper.)  Full explanation of this observation could be interesting future work. 

\clearpage

\section{Autoencoders architecture}

We use the same hyperparameters for each autoencoder and if the representation size allows, the same architecture, even between datasets and models. Each encoder has 3 convolutional layers with 128 filters, the stride set to 2, ReLU activations and a single fully-connected layer with the latent size set to 64. The architectures of the decoder and the encoder are symmetric.

\end{document}